\documentclass[11pt,a4paper]{article}
\usepackage[hyperref]{acl2019}
\usepackage{times}
\usepackage{subfigure}
\usepackage{latexsym}
\usepackage{graphicx}
\usepackage{multirow}
\usepackage{url}

\aclfinalcopy 

\title{Data mining Mandarin tone contour shapes}

\author{Shuo Zhang \\
  CED Applied Research\\
  Bose Corporation \\
  The Mountain Rd, Framingham, MA 01701\\
  {\tt shuo\_zhang@bose.com}}

\date{}

\begin{document}
\maketitle
\begin{abstract}

In spontaneous speech, Mandarin tones that belong to the same tone category may exhibit many different contour shapes. We explore the use of data mining and NLP techniques for understanding the variability of tones in a large corpus of Mandarin newscast speech. First, we adapt a graph-based approach to characterize the clusters (fuzzy types) of tone contour shapes observed in each tone $n$-gram category. Second, we show correlations between these realized contour shape types and a bag of automatically extracted linguistic features. We discuss the implications of the current study within the context of phonological and information theory.


\end{abstract}

\section{Introduction}
\begin{figure}

\small

 \centerline{
 \includegraphics[scale=0.1]{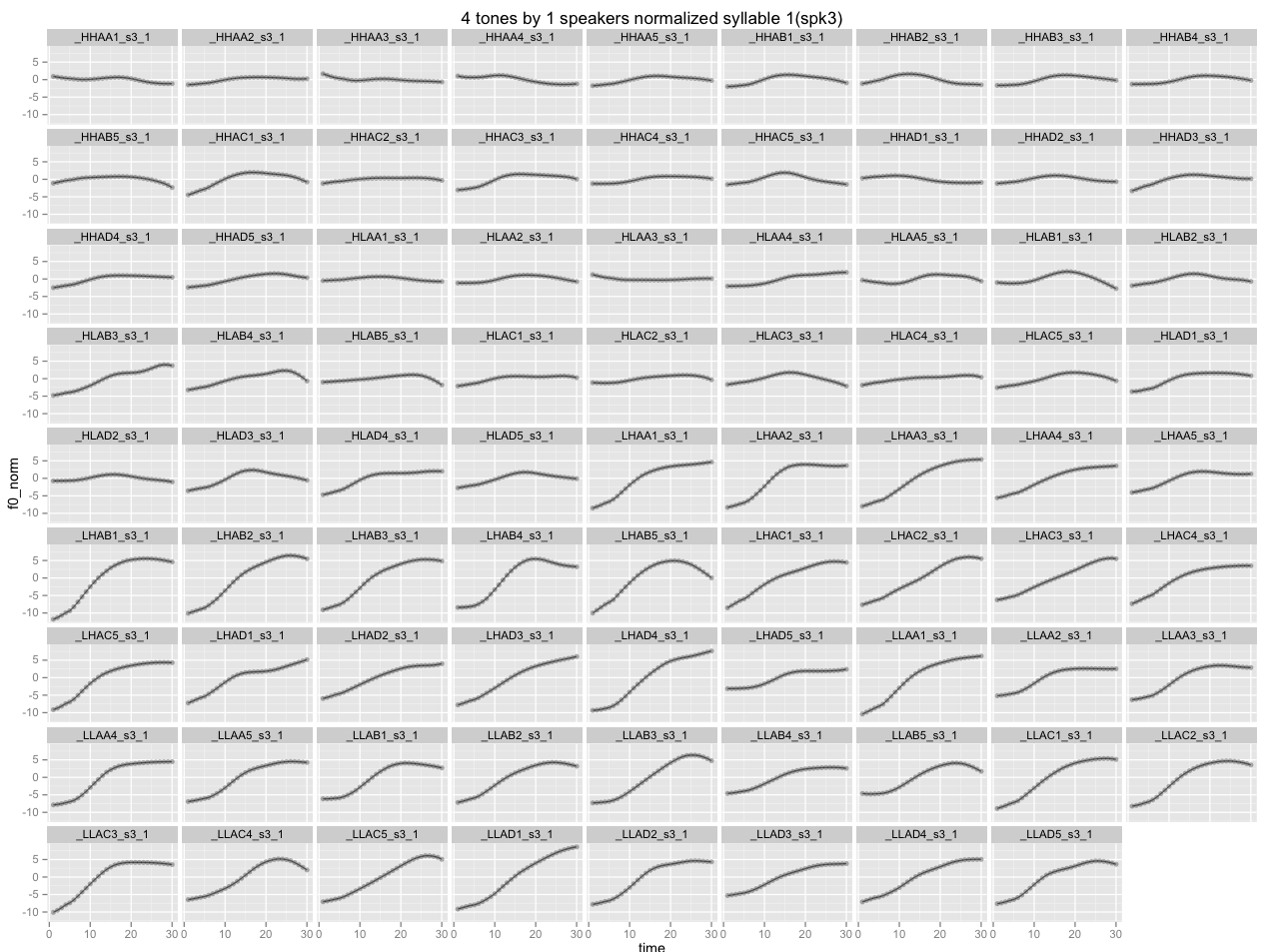}}
 \caption[Samples of Mandarin Tone1 by one speaker in lab speech]{Samples of Mandarin Tone1 by the same speaker in lab speech. Data source: \cite{Xu1997}. The canonical contours of Mandarin Tone 1,2,3,4 are: high level, low rising, low dipping, high falling, where low and high denote the pitch starting point of the tone. }
  \label{fig:tone1}

\end{figure}

One of the central phenomena of interest in lexical tone production is the deviation of their surface realizations from canonical templates of tone categories\cite{Xu1997, Prom-on2009, Surendran2007}. In a tone language, different tone categories differing in pitch movements can distinguish different lexical meanings of a syllable (e.g., in Mandarin, the syllable ``ma" in a high level pitch contour means ``mother", whereas the same syllable spoken in a falling pitch contour means ``to scold"). Even though each tone category is defined with a general pitch contour profile (such as level, rising, falling, etc.), they typically exhibit great variability in spontaneous speech. As an example, Figure \ref{fig:tone1} shows many different realizations of Mandarin tone 1, observed during speech production experiments in the lab. 


Previous works in phonology, speech prosody, and tone recognition have investigated this variability by asking questions such as: (1) What factors contribute to the variability in tone production \cite{Xu1997}?(2) How can we model the tone contour trajectory in synthesized speech \cite{Prom-on2009}? (3) What features can we use to improve the accuracy of automatic tone recognition \cite{Surendran2007}? Each of the works was driven by a particular set of theoretical or practical motivations and offered us a slice of understanding into the problem.

In this work, we are interested in looking at the tone variability problem from a data mining perspective: we explore the structure and distribution of tone contour shapes within a large amount of data. By taking a data mining approach, we contrast our work with those works that focus on tone recognition or tone learning (either by machine or by human): we seek to extract tone patterns of empirical significance from a large data set of tones from spontaneous speech. 

Working with the MCPST corpus (see Section \ref{sec:data}) of Mandarin newscast speech (about 100,000 tones), we ask two questions: (1) For each tone category, what are the (coarse) types/classes of tone contour shapes we observe in this corpus? (2) For a particular tone category, what linguistic factors caused the same tone to be realized as these different types of shapes? 

Inspired by works in natural language processing (NLP), we further extend these research questions in two directions. First, we extend our investigation of tone categories into a series of $n$ consecutive tones, or tone $n$-grams. $N$-grams is a classic technique in NLP language modeling\footnote{Readers may refer to the classic NLP textbook chapter if needed: https://web.stanford.edu/~jurafsky/slp3/3.pdf.}, whereas in the current context, we study tone $n$-grams due to the importance of context in tone variability \cite{Xu1997}: a tone category maybe realized differently depending on their neighboring tones. What can we learn from data mining tone contour shapes for tone unigrams, bigrams, and trigrams?

Second, to study prosody interface in MCPST data, we use automatic methods (NLP and other) to extract linguistic features from the text, including Named Entity Recognition (NER), Coreference resolution, Part-of-speech (POS) tagging, dependency parsing, and other phonological, morphological and contextual features. In order to find out the importance of these linguistic factors in shaping tone variability, we run the following experiment: given a particular tone (or tone $n$-gram) category, how well can we predict the type of tone contour shape it will take in running speech, using these linguistic features that \textit{exclude} information about the pitch contour $f0$ values?

Previous works showed that many linguistic factors (such as focus, topic, etc.) affect tone production or prosody (see Section \ref{sec:related-work}) . In this work, we extend this to a more comprehensive set of linguistic features, motivated by the information theory account of tone production. We hypothesize that there exists an information content inequality resulting from probability distribution of events in various linguistic domains (phonological, semantic, etc). These inequalities affect speakers' speech production, resulting in gradient variants of tone contour shapes in a given tone category. We investigate the relative importance of these factors in predicting the types of contour shapes any particular tone $n$-gram will take.

The rest of the paper is organized as follows. Section \ref{sec:related-work} discusses relevant previous works. Next we describe the data used in this paper in Section \ref{sec:data}. In order to characterize the types of contour shapes a tone $n$-gram will take, we develop a method to derive clusters of tone contour shape types using network analysis (Section \ref{sec:shape}).  In Section \ref{sec:features}, we discuss feature engineering and feature extraction from various linguistic domains (syntax, morphology, semantics, information structure, etc.). Section \ref{sec:ML} reports machine learning experiments and results on predicting tone contour shape types and the analysis on feature importance. Finally, in Section \ref{sec:discussion} we discuss the implications of this work in the context of information theory and phonological theory of speech and tone production.




\section{Related Work} \label{sec:related-work}



There has been a long line of research on the variability of tone contour shapes as well as interfacing between other linguistic factors and prosody \cite{KLi09,buring2013}. In linguistic research of Mandarin tones, most works have focused on the effect of local tonal context (e.g., neighboring tones and pitch range, such as \cite{Gauthier2007, Xu1997}) and broader context (e.g., focus, topic, information structure, long term $f_0$ variations, such as \cite{Xu2004, Liu2006, Wang2011}). The data in these works usually consisted of a small number of tone observations obtained in speech production experiments in the lab. They have informed later works on improving the performance of supervised or unsupervised tone recognition (\cite{Levow2005, Surendran2007} etc.). Other works such as \cite{Surendran2007} and \cite{KYu11} have shown the importance of signals in speech outside of $f_0$ for tone recognition and learning.

In the PENTA  \cite{Xu1997, Xu2005}  and qTA (quantitative target approximation) models \cite{Prom-on2009}, the surface $f_0$ contour is viewed as the result of asymptotic approximation to an underlying pitch target, which can be a static target (High or Low) or a dynamic target (Rise or Fall). An important contribution of the qTA is that it provides a mathematical model to account for the process of generating of a particular realization of a tone template, defined by a pitch target (with slope and intercept parameters) and the acceleration rate. As such, the specific shape of the contour then would depend on the starting pitch, ending pitch target, and how fast the pitch moves.


A fundamental theoretical question is how should we view the underlying factors that account for the tone surface variability. Previous research exhibits two opposing theories to this question. The first approach \cite{Cooper1985, Cooper1981} postulates a direct link between communicative functions and surface acoustic forms by finding the acoustic correlates of certain communicative functions, such as focus, stress, newness, questions, etc. Such approaches have met criticisms from phonologists \cite{Ladd1996, Liberman1984}, who argue that prosodic meanings are not directly mapped onto acoustic correlates. Instead, intonational meanings should be first mapped onto phonological structures, which is in turn linked to surface acoustic forms through phonetic implementation rules. In this work, we attempt to show a new middle ground between these two theories.


\section{Data} \label{sec:data}

All the data in this work comes from the Mandarin Chinese Phonetic Segmentation and Tone (MCPST) corpus \footnote{https://catalog.ldc.upenn.edu/LDC2015S05}, developed by the Linguistic Data Consortium (LDC). It contains 7,849 Mandarin Chinese newscast speech ``utterances" and their phonetic segmentation and tone labels. Utterances are defined as the time-stamped between-pause units in the transcribed news recordings. We used the auto-correlation algorithm implemented in Praat \footnote{Boersma, Paul and Weenink, David (2019). Praat: doing phonetics by computer [Computer program]. Version 6.0.48, retrieved 17 February 2019 from http://www.praat.org/} for $f_0$ (pitch) estimation from speech audio signal. We obtained $f_0$ pitch contour data for 100,161 syllables. After pre-processing the pitch tracks (e.g., speaker-dependent normalization, $f_0$ outlier detection and removal, pitch interpolation, downsampling), we generate tone unigram, bigram and trigram $f_0$ data sets, giving rise to a total of 75 unigram (5), bigram (16), and trigram data sets for the prediction task (54) \footnote{Mandarin Chinese has four regular tone categories plus one neutral tone. Since neutral tones occurs infrequently, we did not include them in the analysis of tone bigrams and trigrams due to data sparsity in the conditional distributions. Similarly, we also excluded ngrams categories where the data points are sparse.}. The total number of tone $n$-grams in these data sets are on the order of 250k. All tone unigram, bigram, and trigram $f_0$ vectors are downsampled to length of 30, 100, and 200 samples respectively.

\section{Deriving tone $n$-gram contour shape types through network analysis} \label{sec:shape}
\subsection{Problem formulation}

We define a tone $n$-gram category as a consecutive sequence of $n$ tones $t_i$, for $i=1,...,n$, where $t_i$ $\in$ $\{0,1,2,3,4\}$, the five tone categories of Mandarin. In this paper we restrict $n$ to $\{1,2,3\}$. Given the set $S$ (represented as a network) that contains all observations of $f_0$ vectors that belong to a particular tone $n$-gram category, an algorithm $A$, defined in this section, partitions $S$ into $k$ clusters, $c_1,c_2,...,c_k$, where all tone contours within $c_i$ are highly similar to each other, and members of $c_i$ maximally distinct from $c_j$ for $i \neq j$. For a particular tone $n$-gram category, we define the centroid $f_0$ vector of $c_i$ to be its tone contour shape type $t_i$. 

If we denote $C$ to be set of types \{$c_1,c_2,...,c_k$\}, our goal in this section is to describe the algorithm $A$ that learns a function $g: S \rightarrow C$. We adapt an algorithm first proposed by  \cite{Gulati16}, which has been shown to be effective in identifying clusters in time-series data such as pitch contours. It also has several advantages over baseline algorithms such as $k$-means clustering, including outlier pruning and no need to determine the number of clusters before hand. 



On a high level, this method represents all tone contours in a data set as a fully connected network $G$. It then filters $G$ using heuristics based on the pairwise similarity of tone contour shapes. After the filtering step, only those nodes that have a similarity score beyond a threshold will remain connected. It then leverages network community detection algorithms to optimize the community structure, effectively deriving tone contour shape types $T$ = \{$t_1,t_2,...,t_k$\}.

\subsection{Network construction}
To construct the network as described above, we first partition all data in MCPST corpus by their tone $n$-gram category. For each category, we construct a network where each node stores the $f_0$ vector of an observation, and the edge between two nodes holds the Euclidean distance between the two $f_0$ vectors as weights. We derive an undirected, weighted and fully connected network $G$ of tone $n$-gram patterns for each $n$-gram category.  

\subsection{Network filtering}
In this step, we take a fully connected network $G$ of a given tone $n$-gram category and use a principled method to remove edges from the network. Our goal is to find an appropriate threshold so that all edges whose weights (distance between two tone $n$-gram $f_0$ vectors) are greater than the threshold will be cut. In the resulting network, only those nodes representing similar enough tone contour shapes will remain connected.

Specifically, we decide the threshold value by a six-step process: (1) We search for the appropriate threshold in the set of values $\Phi \in \{1.0,1.5,2.0,2.5,3.0,3.5,4.0,4.5\}$ for bigrams and trigrams, $\Phi \in \{0.2,0.4,0.6,0.8\}$ for unigrams. These values are empirically chosen; (2) we iterate over this set of values and each time apply a threshold to the network; (3) after we applied the threshold we convert the network to a unweighted network $G'$ where only those nodes that have a distance below the threshold will remain connected; (4) we produce a randomized network $G_r$ by randomly swapping edges from $G'$ $k$ times while keeping the degree of the nodes constant, where $k$ is equal to the number of edges in $G'$. This can be seen as producing a maximally random network given the degree distribution of the current network; (5) we compute the difference in Clustering Coefficient (CC) of both $G'$ and $G_r$; (6) after repeating this for all values in $T$, we pick the threshold that has the largest difference of CCs.  \footnote{Clustering coefficient (CC) measures the extent to which the nodes in a network tend to cluster together. Intuitively, it expresses how saturated the network is --- how many of the possible connections are actually expressed. The CC for a network of $k$ nodes and $n$ edges is computed as:

\[
    CC = \frac{2n}{k(k-1)}
\]

}


\subsection{Community detection}
\begin{figure}

\small

 \centerline{
 \includegraphics[scale=0.56]{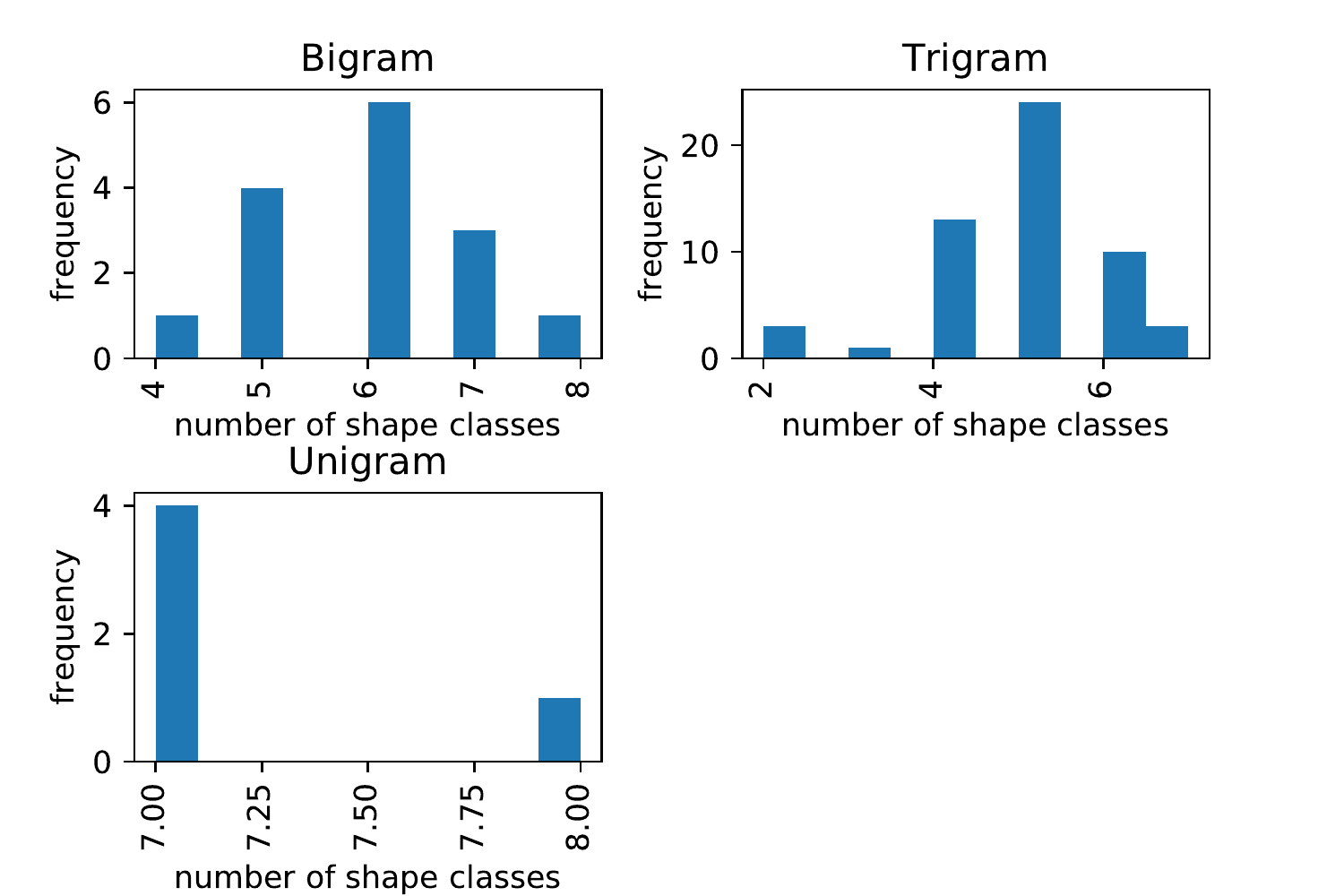}}
 \caption{Histogram of number of shape classes in $n$-gram data sets.}
  \label{fig:num_labels}

\end{figure}

We use the Louvain algorithm \cite{Blondel2008} to perform community detection, in order to partition the filtered network derived from last step into communities (clusters) $C_1, C_2, ..., C_k$. We pre-tuned the hyperparameters in the network filtering step so that it will result in a small number ($n<10$) of tightly connected medium-sized communities. Figure \ref{fig:num_labels} shows a histogram of number of shape classes for unigram, bigram, and trigram data sets.


\subsection{Outlier community filtering}
We propose an extra step of outlier community filtering before deriving our final contour profile classes. In this step, we use a heuristic threshold of $t=10$ to filter out any communities (clusters) with a size less than $t$. 

\subsection{Evaluation of tone contour profile classes}

\begin{figure*}%
\subfigure[Tone 1]{%
  \label{fig:Total_scatter}%
  \includegraphics[width=0.31\textwidth]{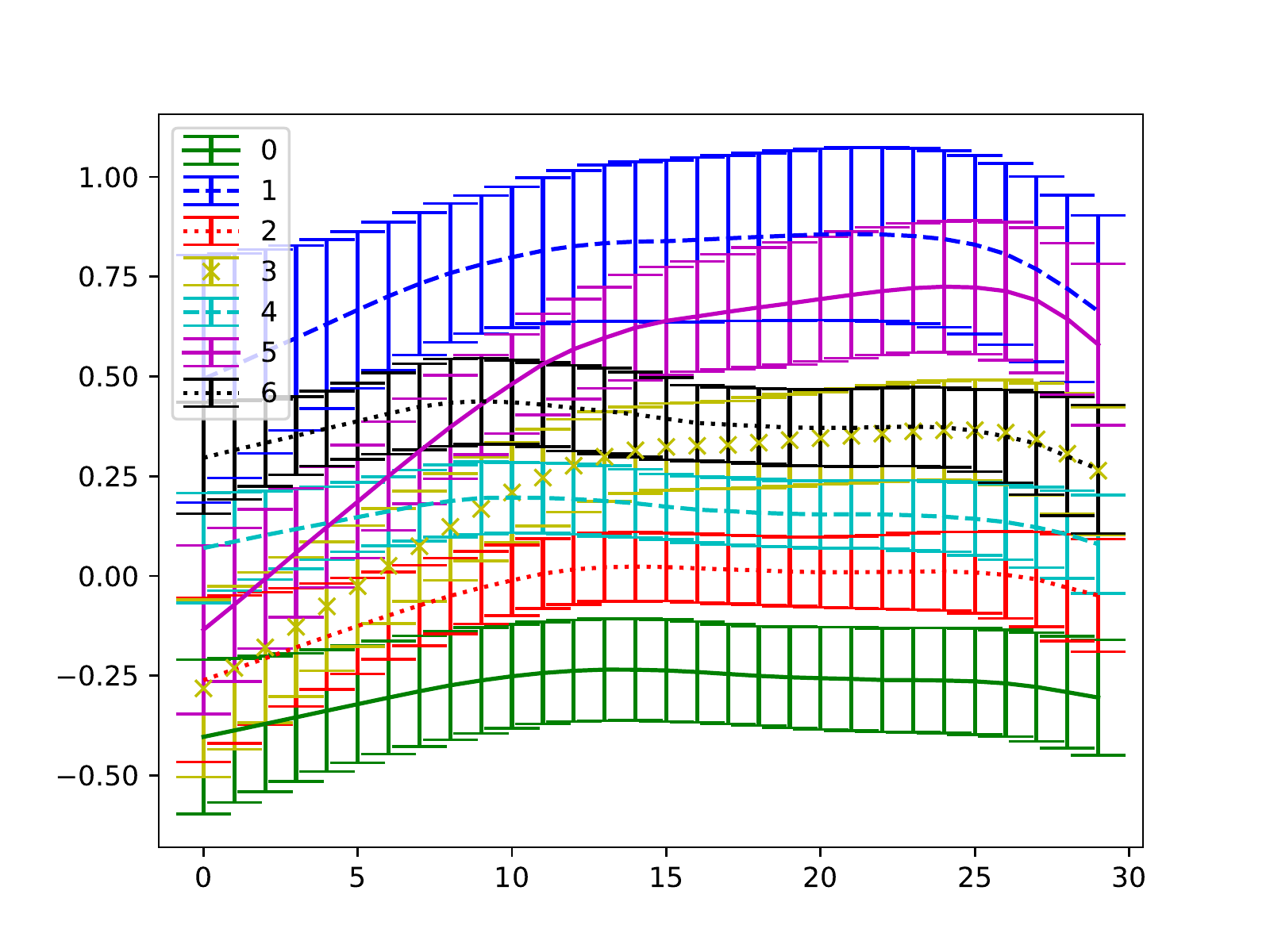}%
}%
\hspace*{\fill}
\subfigure[Tone 2]{
  \label{fig:Num_scatter}%
  \includegraphics[width=0.31\textwidth]{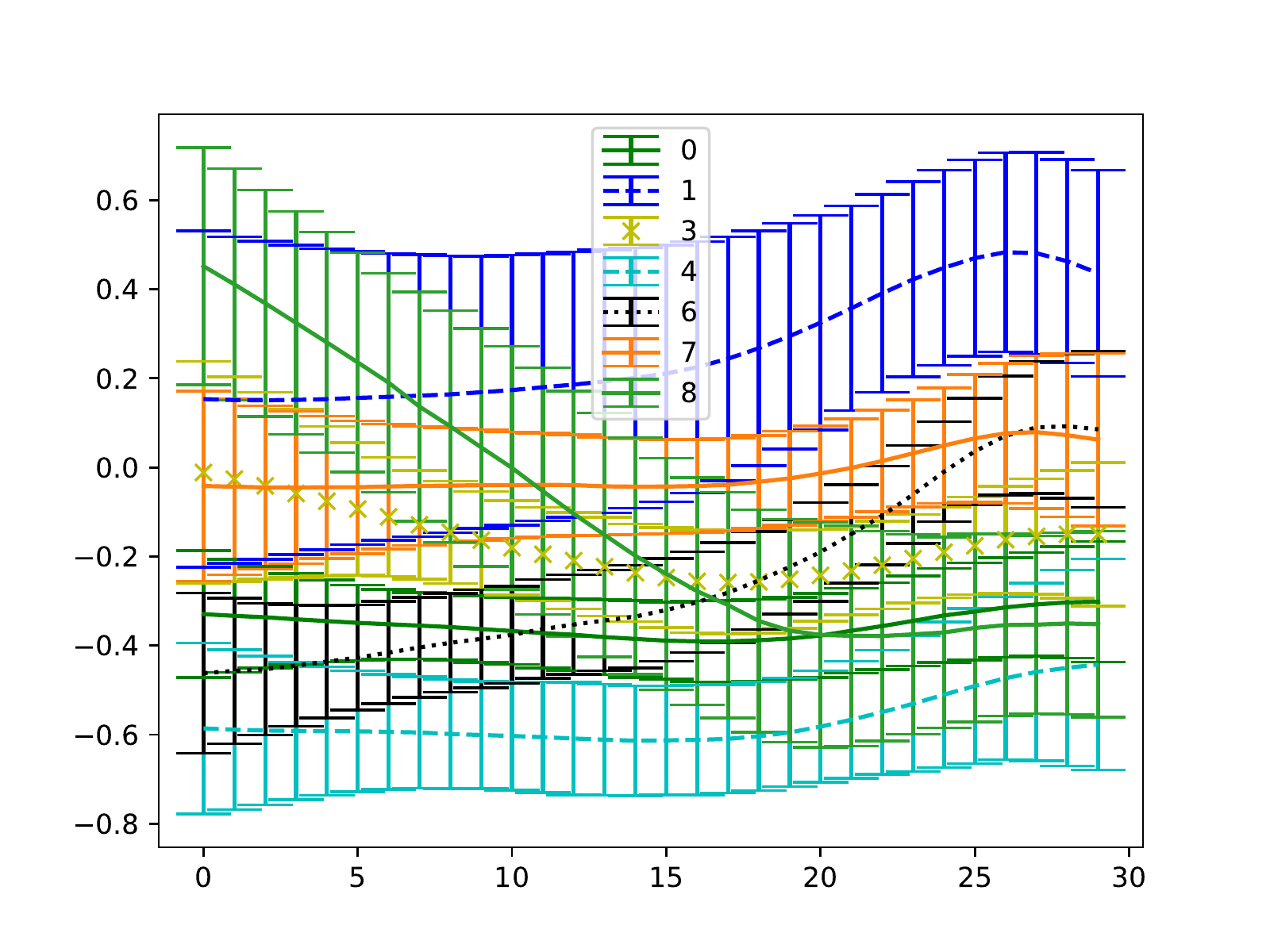}%
}%
\hspace*{\fill}
\subfigure[Tone 2-3]{
  \label{fig:Raw_scatter}%
  \includegraphics[width=0.31\textwidth]{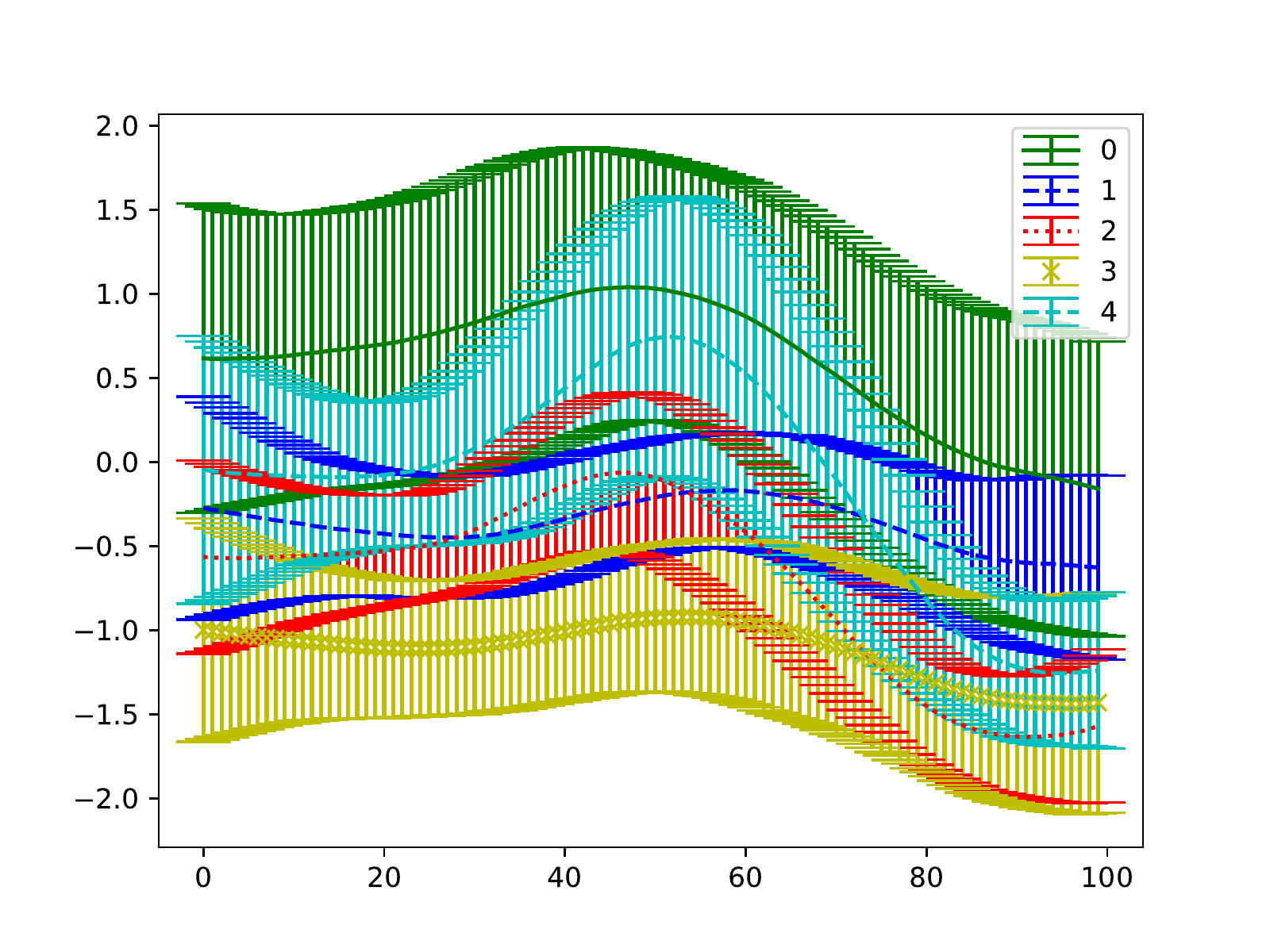}%
}

\subfigure[Tone 4-2]{%
  \label{fig:Num_Raw_scatter}%
  \includegraphics[width=0.31\textwidth]{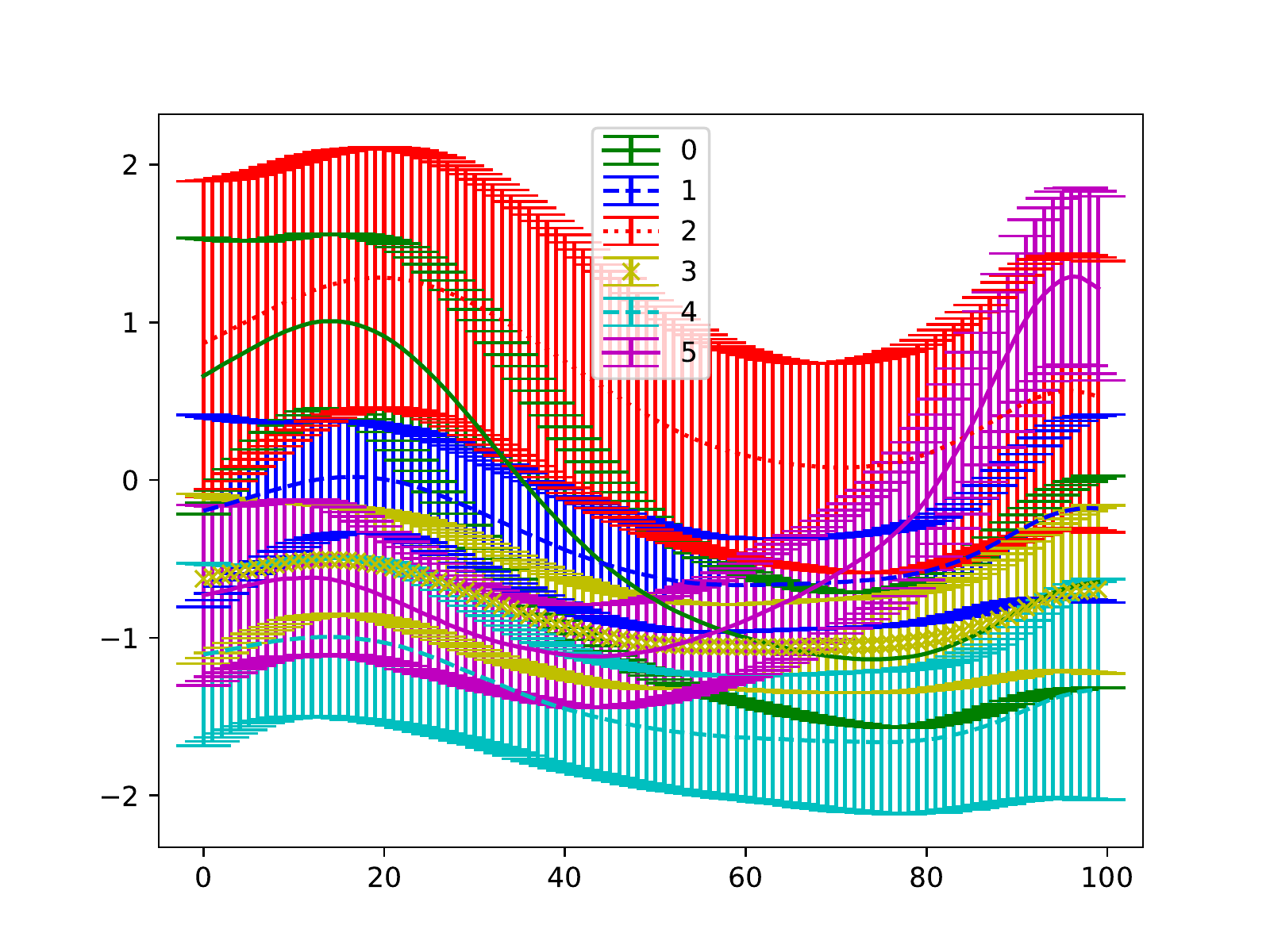}%
}%
\hspace*{\fill}
\subfigure[Tone 3-4-2]{
  \label{fig:Total_Raw_scatter}%
  \includegraphics[width=0.31\textwidth]{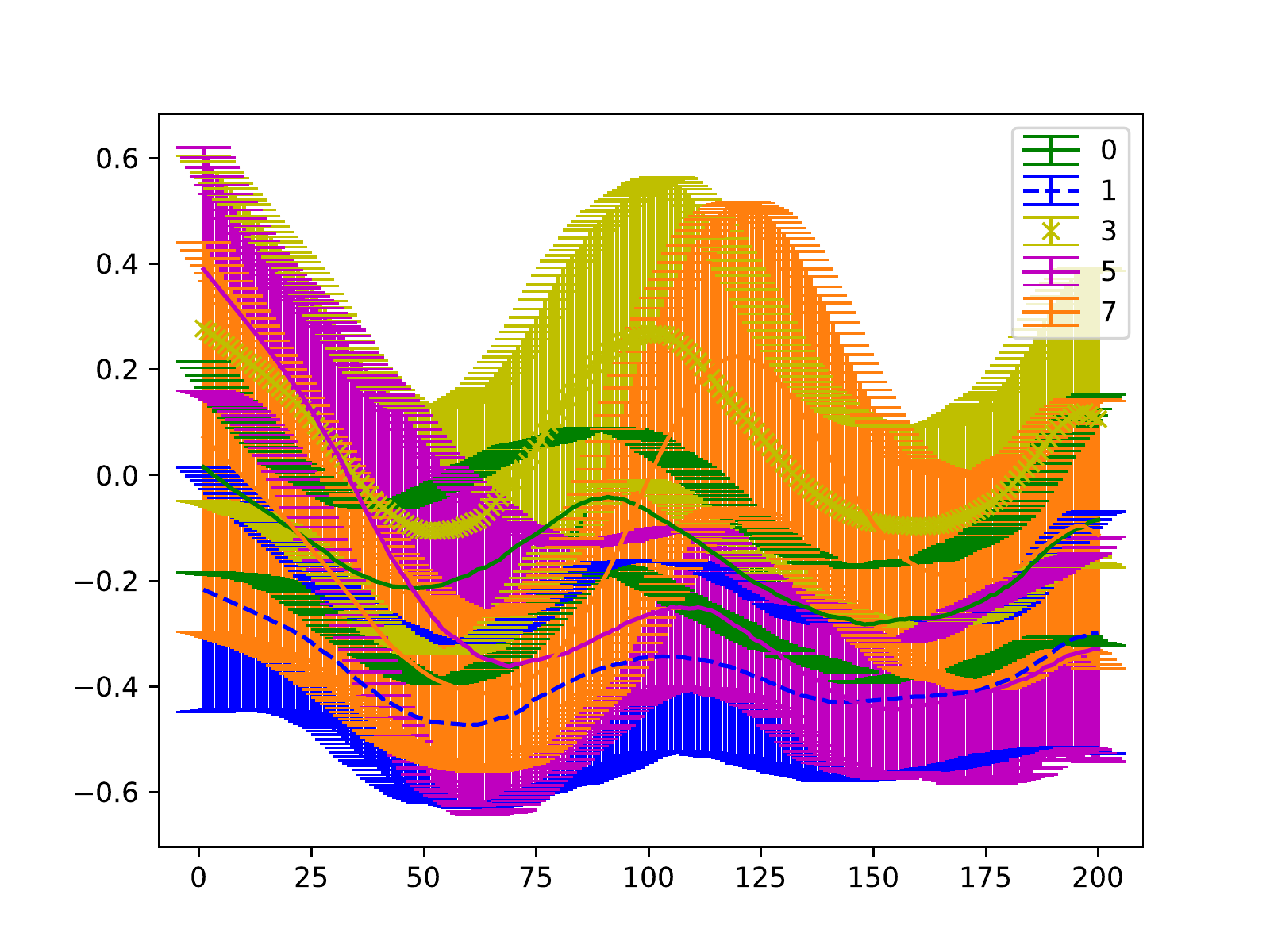}%
}%
\hspace*{\fill}
\subfigure[Tones 2-2-4]{
  \label{fig:Num_Total_scatter}%
  \includegraphics[width=0.31\textwidth]{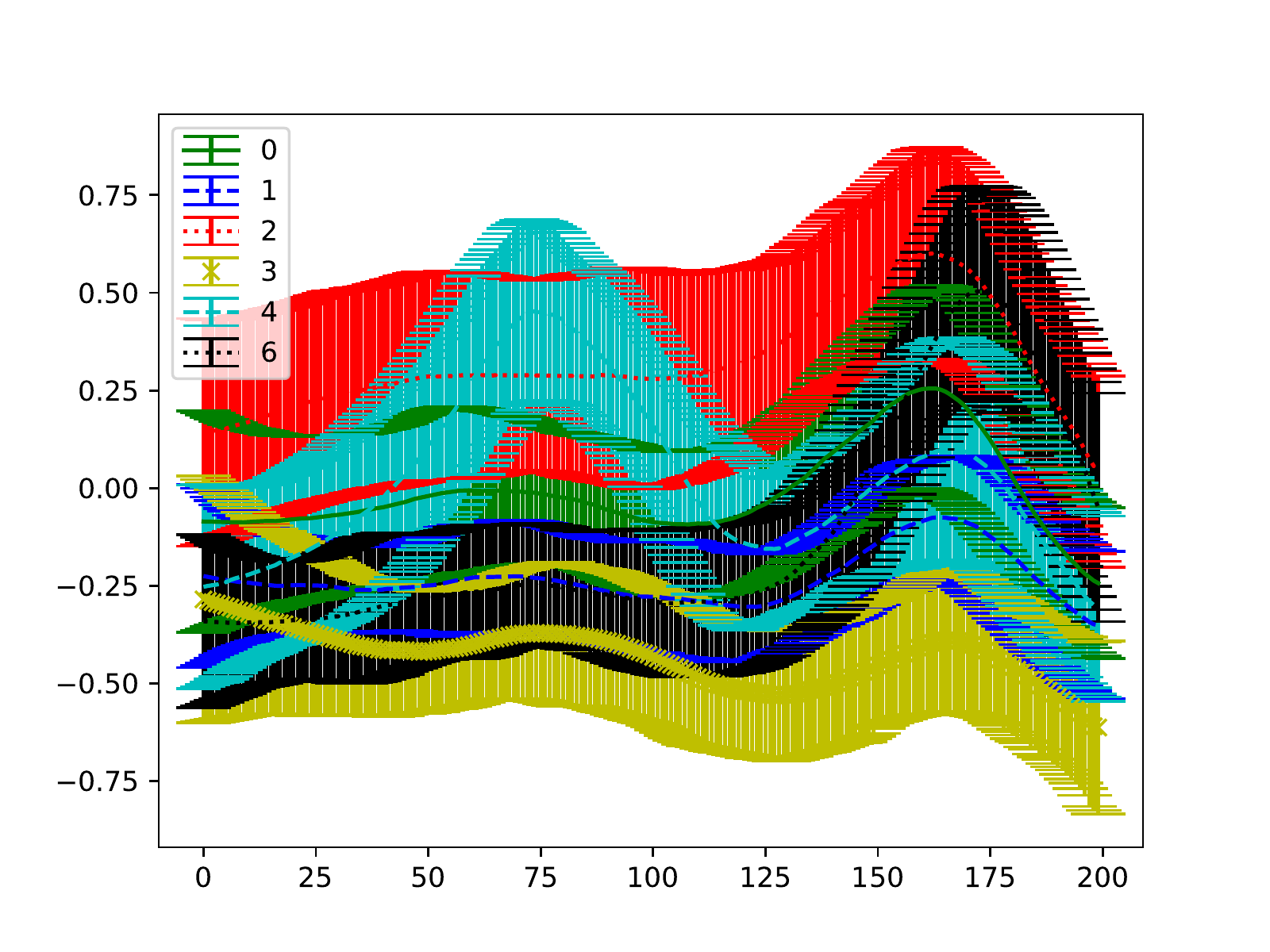}%
}

\caption{Example contour shape type clusters found in randomly selected tone $n$-gram categories, with two examples from each of tone unigram, bigram, and trigram. Each cluster is represented by a mean pitch vector with error bars (as shown in the legend). Clusters are indexed by integers shown in legend. X-axis shows number of samples (discrete time index) for the tone contour $f_0$ vector. Y-axis shows the speaker-normalized pitch values.}\label{fig:panel}
\end{figure*}

We have leveraged the intrinsic structure of the tone data to derive the tone contour shape types for each tone $n$-gram category. Figure \ref{fig:panel} shows examples of learned clusters of tone contour shape types from two $n$-gram categories of tone unigram, bigram, and trigram, respectively. Without declaring any cognitive or phonological significance of these clusters, these resulting clusters should reflect the similarities of tone contour shapes within any given tone $n$-gram category: those that are highly similar are grouped into the same cluster. This is an intrinsic property derived from the above method, and is a necessary property sufficient for carrying out the subsequent experiment on predicting the tone contour shape types from linguistic factors.

Nonetheless, we propose two different ways to evaluate the validity of these clusters. First, in the following experiments, we show that we are able to predict these learned tone contour shape types significantly better than randomly assigned clusters (Section \ref{sec:results}, Figure \ref{fig:acc-box}). Second, we train a decision tree classifier to predict the shape type of a given tone $n$-gram using its $f_0$ vector and obtained a mean accuracy of 92\% (following \cite{W16-2001}). This indicates how well these tone contour shape types can be predicted \textit{with} complete information of its pitch trajectory, which will serve as an upper bound to our next prediction task using linguistic factors \textit{without} information about pitch movements $f_0$ values.



\section{Linguistic features} \label{sec:features}



After we obtained tone contour shape types emerged from each tone $n$-gram category in the corpus, we now describe the linguistic features used to predict which type of shape the tone $n$-gram will take. All syntactic and semantic features are extracted using Stanford CoreNLP for Chinese \cite{manning-EtAl:2014:P14-5}. 


\subsection{Syntactic features}
Syntax and prosody has been the subject of investigation in (psycho)linguistic studies  \cite{Bard99thedissociation}.  We extract the part-of-speech (POS) tags for all syllables in a tone $n$-gram. In addition, we also extract the dependency function \cite{DBLP:conf/emnlp/ChenM14} of all syllables in the tone $n$-gram. Therefore there are $2*N$ syntactic features where N is the number of syllables included in the tone $n$-gram data under consideration. The original tag set used in CoreNLP comes from Penn Chinese Treebank\footnote{https://catalog.ldc.upenn.edu/LDC2013T21} and is too fine grained. To avoid data sparseness, we collapsed several categories. For both POS tags and dependency edge function categories, we compute their distributions using the original tag set and we collapse any categories that appear less than 5 times in the data. For POS tags we mapped the original 33 tags onto 5 categories. For dependency functions, we collapsed all tags with a subcategory separated by a colon (e.g., ``advmod:loc'', ``advmod:rcomp'', mapped to ``advmod'' etc.). 

\subsection{Semantic features}
We extract two semantic features for a tone $n$-gram data point: (1) whether the tone $n$-gram includes a named entity; (2) whether the tone $n$-gram includes a singleton (as opposed to being part of a coreference chain in the discourse). Semantic features such as information structure have been postulated to have an effect on the prosody domain \cite{buring2013}. In particular, given information may encode prosodic features different from new information. This could also apply to named entities vs. non-named entities. Named entities points to definite, specific objects in the real world. Whether the token is a singleton (i.e., does not co-refer to an entity with another mention in the text) or part of a coreference chain can be correlated with information structure \cite{recasens-etal-2013-life}. That is, a singleton may signify new information in discourse, while a non-singleton is part of a coreference chain with potential antecedent or anaphor, pointing to potentially a different information structure. Both can have distinguishing effects on the mental representations and the production of speech prosody (indirectly related to redundancy in \cite{Aylett2004}).

\begin{table*}
 \caption[Feature set overview]{Feature set overview. \textmd{1...N indicates this feature is computed for all syllables in the $N$-gram. Total number is N*10+7 features, 37 for trigrams and 27 for bigrams, etc.}}
\small

\begin{center}
\begin{tabular}{ |c|c|c|c|c| } 

 \hline
Syntactic & Morphological & Semantic & Phonological & Others\\
 \hline
\hline
$POS\_Tag_{1...N}$ & $Tok\_Bound_{1...N}$ & is\_entity & $is\_nasal_{1...N}$ & sent\_position \\ 
$Dep\_Func_{1...N}$ & & is\_singleton & $is\_dipthong_{1...N}$ & start\_pitch \\
 & & & $is\_round_{1...N}$ & end\_pitch\\
 & & & $is\_front_{1...N}$ & prev\_tone\\
  & & & $is\_back_{1...N}$ & next\_tone\\
   & & & $is\_high_{1...N}$ & \\
    & & & $is\_low_{1...N}$ & \\

\hline
\end{tabular}

  \label{tab:trigramFeatures}
\end{center}
\end{table*}

\subsection{Morphological features}
In Mandarin Chinese, each word usually consists of one to four syllables. Building on the intuition that the first syllable is usually spoken with higher prominence (e.g., neutral tone, which does not carry stress, only occurs on word-final positions), we extract morphological features for each syllable in the given tone $n$-gram: whether they cross word boundary or not. There are $n$ features in this category in total.

\subsection{Phonological features}
A basic representation of phonological features is the identity of phonemes in each syllable of the $n$-gram. However, due to the sparseness of this feature representation, we have designed 7 binary features to encode the phonological properties of the syllables in the tone $n$-gram: (1) whether the syllable includes a nasal; (2)  whether the syllable includes a dipthong; (3) whether the syllable includes a high vowel; (4)  whether the syllable includes a low vowel; (5)  whether the syllable includes a front vowel; (6)  whether the syllable includes a back vowel; (7)  whether the syllable includes a round vowel. In addition, we add two contextual tone features: the tone identity of the previous and following syllables of the tone $n$-gram in question.

\subsection{Other features}
We add two pitch features to the feature set: the beginning and ending pitch of the tone $n$-gram. This is based on the notion in generative Mandarin tone modeling (Parallel ENcoding and Target Approximation model, or PENTA) that in speech production, the actual realized tone shape of a given tone category highly depends on the starting point of the pitch contour and its distance to the actual pitch target of the current tone, which affects its course of trajectory when it approximates the target \cite{Prom-on2009}. An additional feature to be included is the position of the current tone $n$-gram within the context of the current sentence as a percentage. It is a known effect that pitch tends to downdrift in speech production as sentence progresses \cite{Wang2011}. Therefore, we also want to account for the effect of sentence position.

\subsection{Bag of features}
In this task, we note that the unit of feature extraction is not as straightforward as it would be in classic NLP tasks. That is, instead of a typical syntactic constituent (word, phrase, sentences) as the feature extraction unit, here, our target is tone $n$-gram, a sequence of $n$ syllables that may or may not be a syntactic constituent. As described above, in many features we have adopted a  ``Bag of features'' approach (similar to the speech coreference resolution work in Roesiger et al.  \cite{Roesiger2015}) where each feature describes whether the $n$-gram contains a certain target value in any position. For some other features, we simply use a set of $n$ features applied to each syllable in the $n$-gram in question. These are precisely described in Table \ref{tab:trigramFeatures}.


\section{Predicting tone contour profiles} \label{sec:ML}

\begin{figure}

\small

 \centerline{
 \includegraphics[scale=0.58]{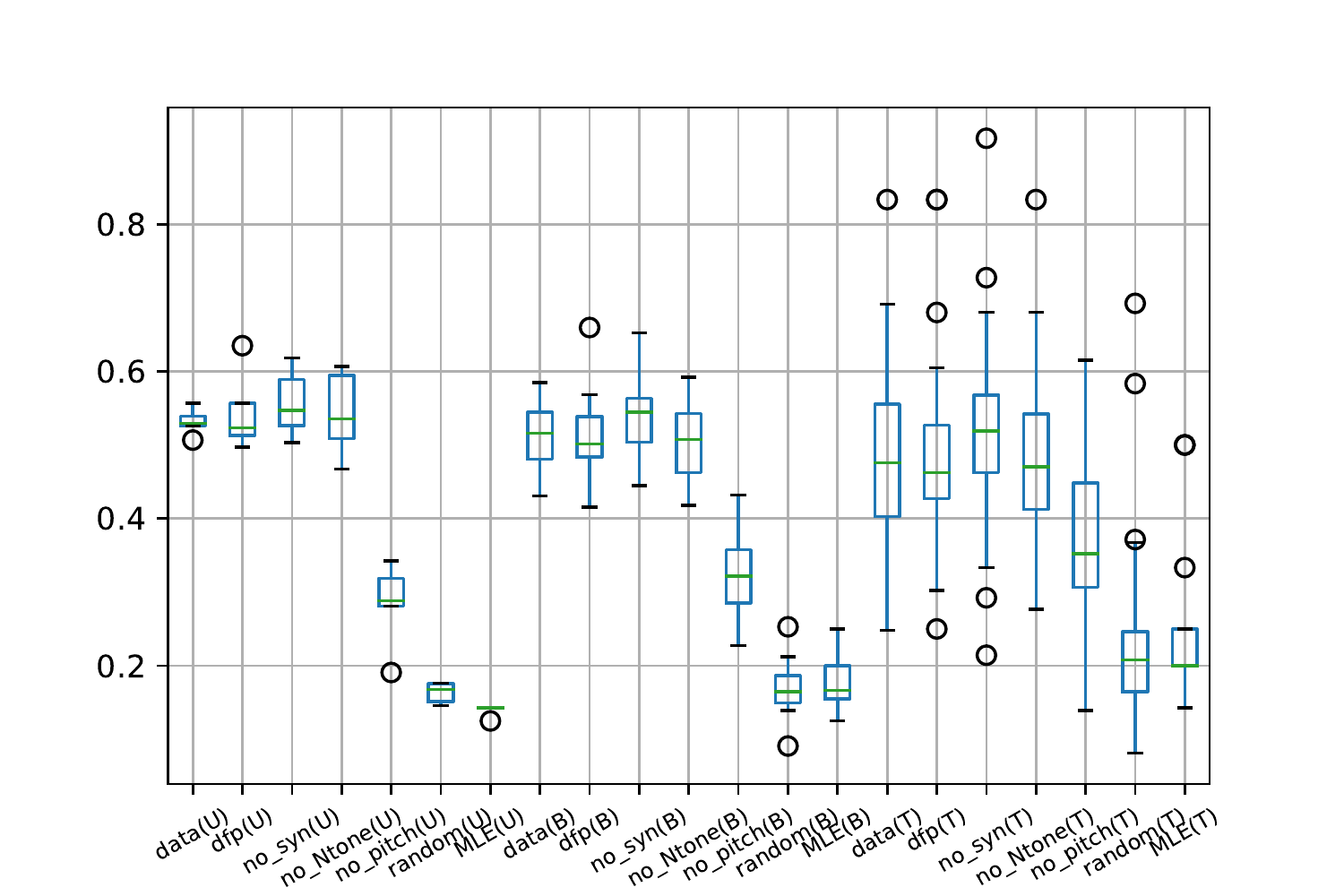}}
 \caption{Distribution of classification accuracies for all 75 data sets in unigram(U), bigram(B), trigram(T). For each $n$-gram the labels on the x-axis are: data:full feature set; dfp:start/end pitch only baseline; no\_syn:without syntactic features; no\_Ntone:without prev/next tone features; no\_pitch:without start/end pitch features; random:baseline with randomly assigned labels; MLE:MLE baseline.}
  \label{fig:acc-box}

\end{figure}

\subsection{Experimental setup}

For each of the 5 unigram, 16 bigram, and 54 trigram data sets, the extracted linguistic feature vector ($f_1,f_2,...,f_m$) forms the input space $X$. The contour shape types $T$ forms the output space $Y$. Our goal is to learn a function $h: X \rightarrow Y$ minimizing expected loss. We use SVM with linear kernel so that we can extract feature importance in subsequent analyses. Each data set is randomly split into 90\slash10 for train and test. Since the classes are balanced in $Y$, we evaluate the classifier performance directly with accuracy on the test sets. 

\subsection{Results} \label{sec:results}
To visualize model performances on a large number of tone $n$-gram data sets (75), we choose the boxplot because of its efficiency to convey statistical information of the distribution of the results across all data sets. Figure \ref{fig:acc-box} gives an overview of our proposed model classification accuracies for all unigram(U), bigram(B), and trigram(T) data sets, as compared to several baselines. In this figure, \texttt{data}, \texttt{dfp}, \texttt{no\_syn}, \texttt{no\_Ntone}, and \texttt{no\_pitch} denote results using different sets of features: full set, start/end pitch only, no syntactic features, no prev/next tone features, and no start/end pitch features, respectively. The \texttt{random} baseline uses the same set of linguistic features as our proposed model but the target tone contour shape type is randomly assigned (while keeping the number of shape types constant). Finally, the Maximum Likelihood Estimation (\texttt{MLE}) baseline reflects chance level performance if linguistic factors are independent from the output tone contour shape types, and is calculated as $1/d$, where $d$ is the number of output classes in a data set.

First, the proposed model significantly outperforms MLE baselines for unigrams, bigrams, and trigrams. Second, the accuracy for predicting learned tone contour shape types is significantly higher than randomly assigned clusters. This serves as a sanity check for the validity of these learned tone contour shape types. Overall, this result supports the hypothesis that a variety of linguistic and contextual features are strongly correlated with the realization of a particular category of tone $n$-grams. In particular, we observe that using the set of features excluding the syntactic features (POS tags and dependency functions) allows the model to achieve the best median accuracy in all $n$-grams (\texttt{no\_syn} baseline). In Section \ref{sec:feature-imp}, we provide a more detailed analysis and discussion of feature importance.

\begin{figure}

\small

 \centerline{
 \includegraphics[scale=0.56]{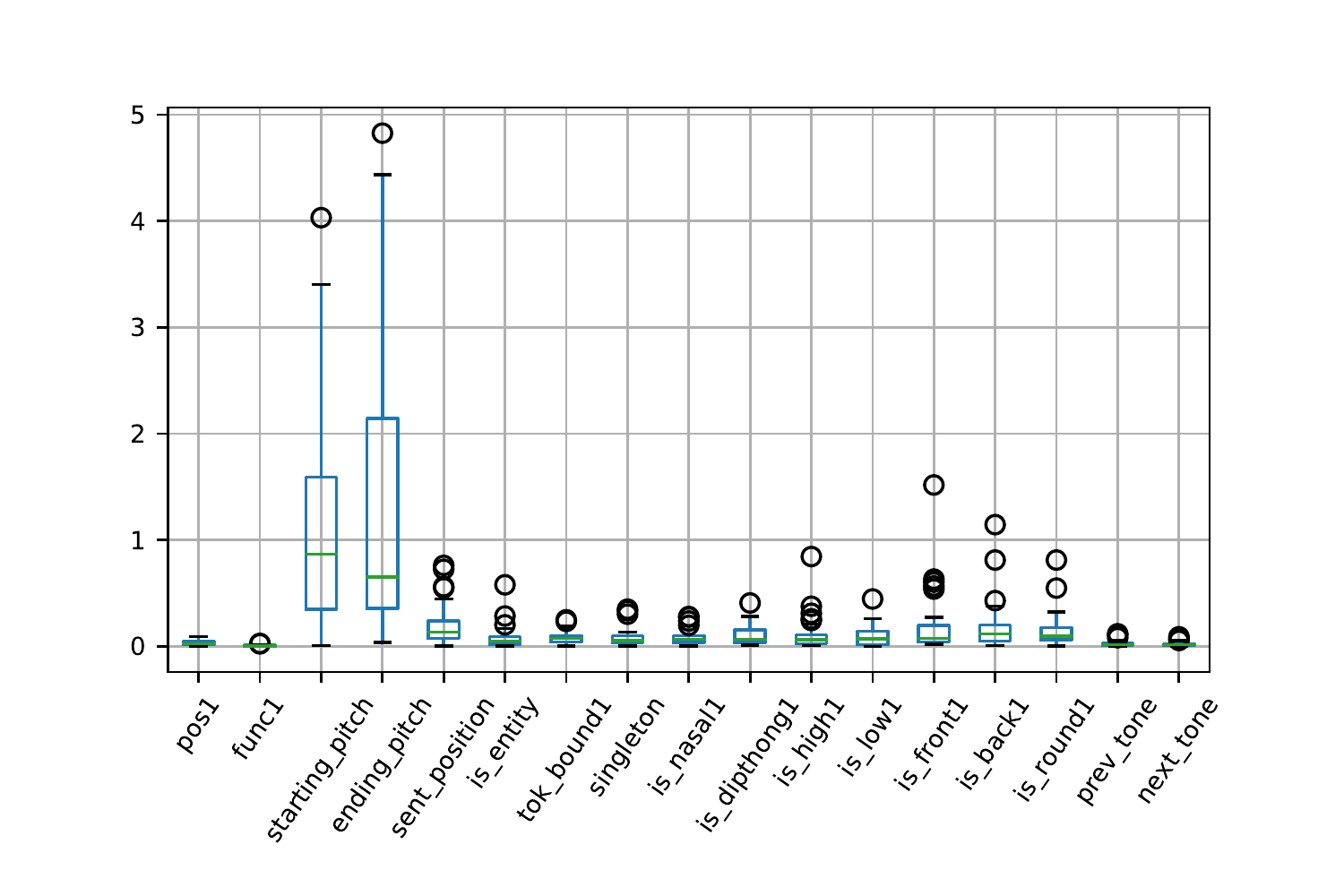}}
 \caption[Feature weights for all unigram data sets]{Feature weights for all unigram data sets.}
  \label{fig:unigramcoef}

\end{figure}

A significant trend to note in Figure \ref{fig:acc-box} is that we observe a negative correlation between the model accuracy and the baseline (MLE, random) accuracy as the value of $n$ becomes larger in $n$-grams. This is striking because it indicates a decrease in predictive power as $n$ grows larger. Moreover, the variance on model accuracy also increases as $n$ becomes larger. From inspecting the tone contour shape types we obtained (such as those showed in Figure \ref{fig:panel}), we attribute this to three factors from the data perspective: (1) the dimensionality of unigram, bigram, and trigram $f_0$ vectors in our data set are different (increase); (2) the number of data sets also increases as $n$ grew larger; (3) the complexity of tone contour shapes tend to increase from unigram to bigram to trigram.  On a linguistic level, we hypothesize that the longer the window of $n$-grams, the stronger an effect of unaccounted factors come into play (longer range prosodic factors such as focus and topic, as demonstrated in \cite{Xu2004}).


\subsection{Feature importance} \label{sec:feature-imp}


To analyze feature importance, we extracted weight vectors (coefficients) associated with all features in the linear SVM classifiers (following \cite{Guyon2002}) for all data sets in each of the $n$-grams. We take the absolute values of feature weights and normalize them to be comparable across data sets. We then aggregated feature weights for all classes and across all data sets in a given $n$-gram. 

\begin{figure}

\small

 \centerline{
 \includegraphics[scale=0.56]{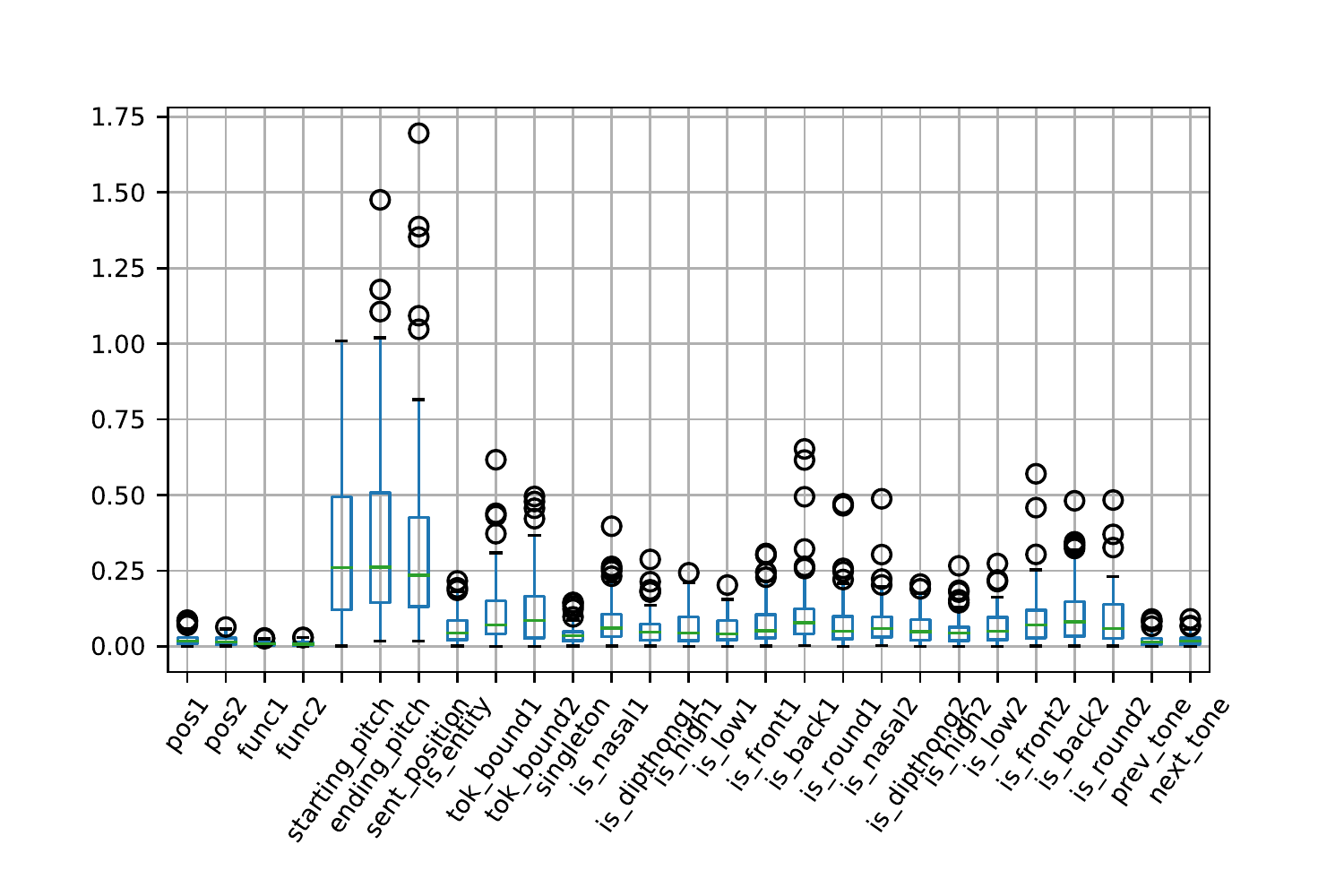}}
 \caption[Feature weights for all bigram data sets]{Feature weights for all bigram data sets.}
  \label{fig:bigramcoef}

\end{figure}
\begin{figure}

\small

 \centerline{
 \includegraphics[scale=0.56]{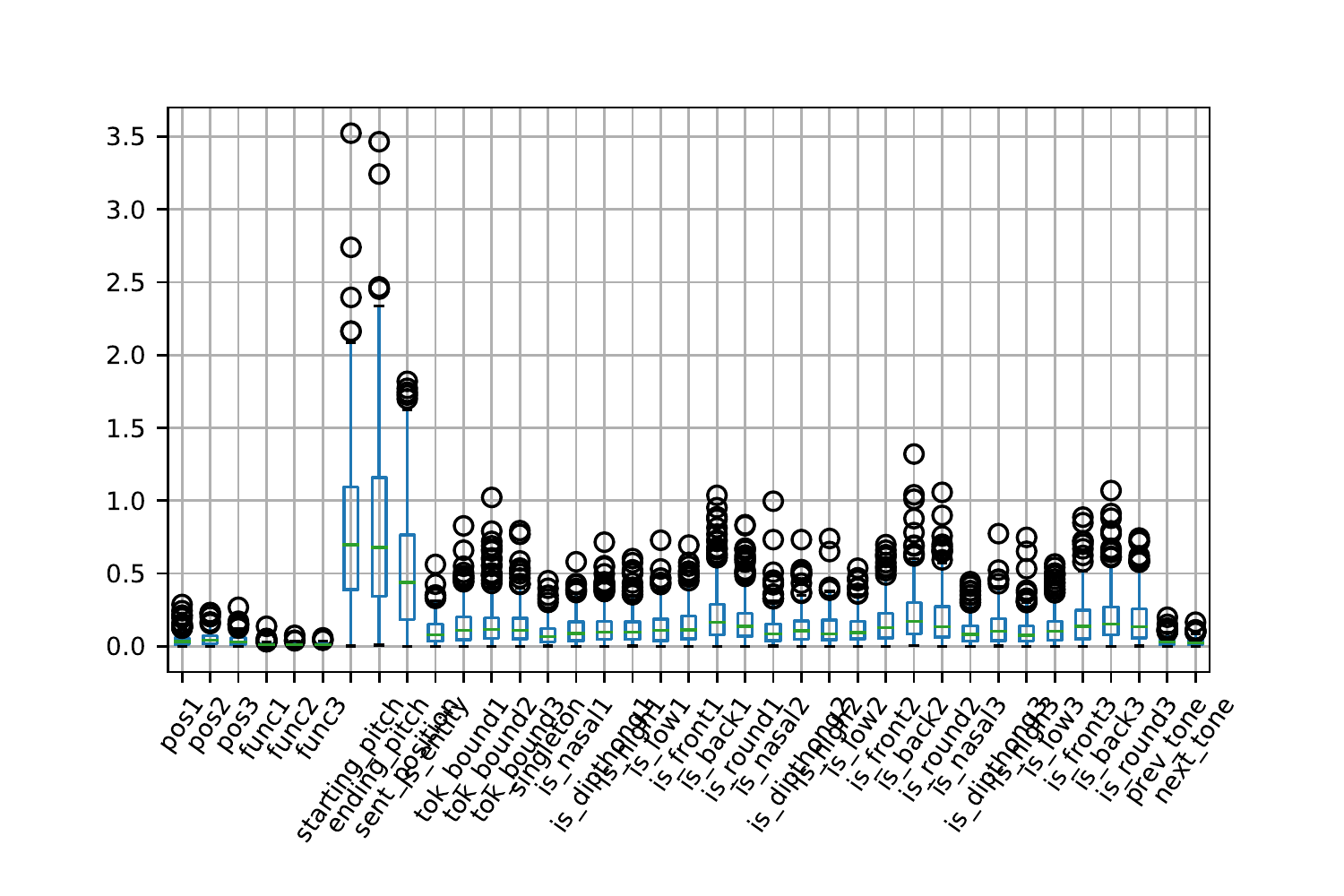}}
 \caption[Feature weights for all trigram data sets]{Feature weights for all trigram data sets.}
  \label{fig:trigramcoef}

\end{figure}

Figure \ref{fig:unigramcoef}, \ref{fig:bigramcoef}, \ref{fig:trigramcoef} show the distribution of feature weights (importance) for all features for unigram, brigram, and trigram data. Since feature importance values tend to behave similarly according to their linguistic domains, we group them together instead of analyzing the features individually. We observe three levels of feature importance based on the weights, consistent among different $n$-grams: (1) High: starting pitch, ending pitch, and sentence position (especially when $n>1$). The importance of starting and ending pitch is consistent with the qTA model of Mandarin tones \cite{Prom-on2009}. The latter (sentence position) is consistent with the effect of downdrift \cite{Wang2011}. (2) Medium: Phonological features and morphological token boundary features, as well as coreference (singleton) and entity, which are information structure of the discourse and semantics (givenness, newness of information in speech). (3) Low: Syntactic (pos tag, dependency functions) and contextual features (previous and next tone). This is consistent with \cite{Bard99thedissociation} and \cite{Surendran2007}\footnote{Specifically, \cite{Bard99thedissociation} showed the dissociation of syntax with de-accent in spontaneous speech, and  \cite{Surendran2007} showed that context did not help tone recognition as much as expected.}. 

\begin{figure}

\small

 \centerline{
 \includegraphics[scale=0.56]{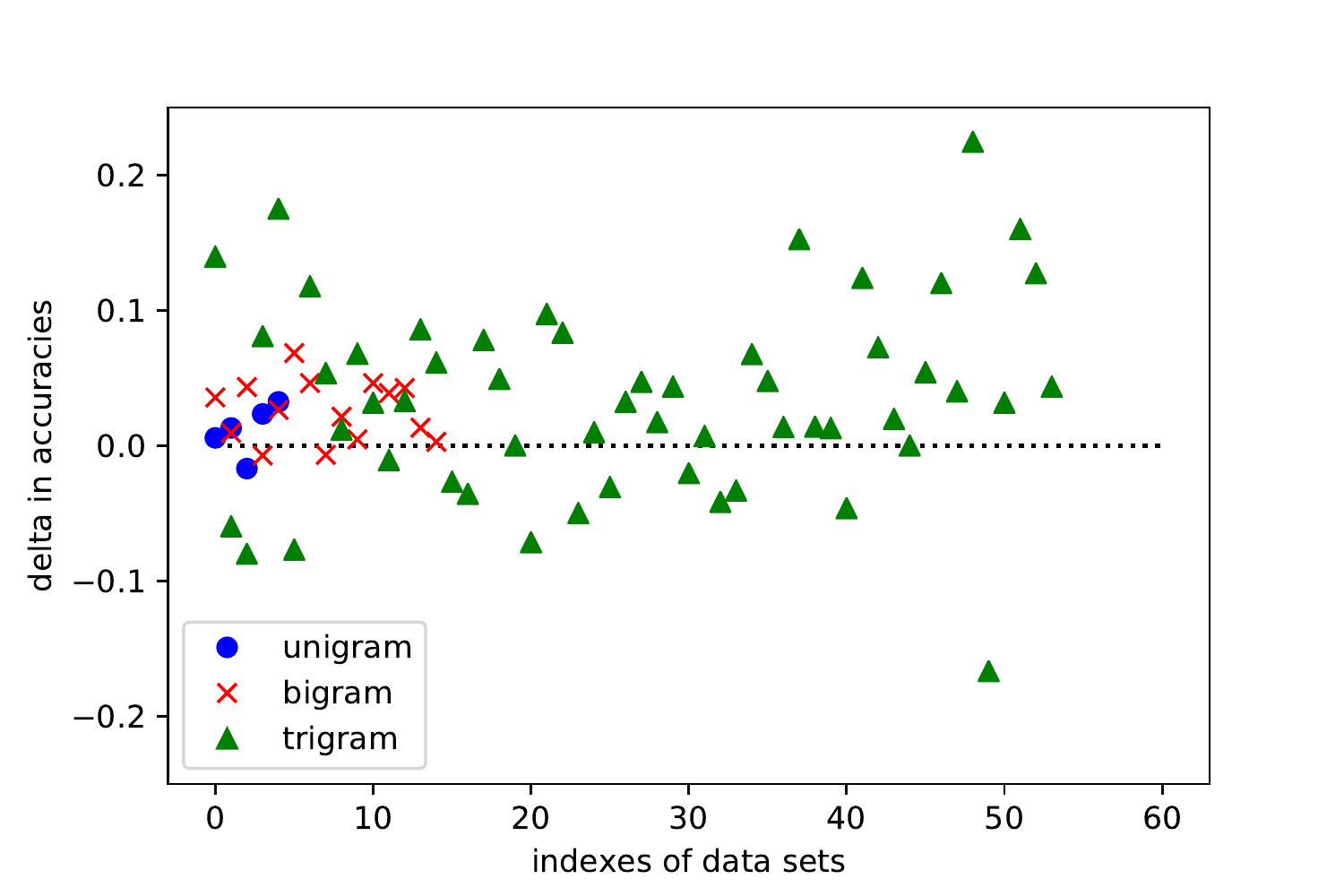}}
 \caption{Delta (differences in accuracies): no\_syn - dfp baselines. A point above the y=0 line indicates the model with linguistic features outperforms the model with pitch only features. }
  \label{fig:diffs}

\end{figure}

To have a more detailed understanding of the feature importance, Figure \ref{fig:acc-box} shows how the model performs with partially ablated feature sets across different $n$-grams. First, to understand the role of start/end pitch vs. non-pitch linguistic features, we observe that the \texttt{dfp} baseline (using only start/end pitch features) has lower results than other models with linguistic features. This is true for all the markers on the boxplots (min, max, first quantile, median, third quantile) when comparing \texttt{dfp} to \texttt{no\_syn}. However, there is also considerable overlap in these accuracy distributions to different degrees as $n$ varies, which indicates cases where the \texttt{dfp} baseline outperforms the other models with more linguistic features. To see this possibility, we plotted the difference (delta) in accuracies values for all data sets for the pair of baselines \texttt{no\_syn} - \texttt{dfp} in Figure \ref{fig:diffs}. It shows that for 80\% of the data sets, the \texttt{no\_syn} baseline with linguistic features outperforms the pitch-only baseline in most data sets by a margin of less than 20\% in accuracy improvements.

Second, comparing across different $n$-grams, the \texttt{no\_pitch} baseline (only linguistic features) performs worse in unigram, and the best in trigrams. This shows that the pitch feature is less important when $n$ becomes larger in $n$-grams. The same trend is observed in the weights of the pitch features in feature importance. This observation is also consistent with \cite{Prom-on2009}: the target approximation of tones only operates on the syllable units. Therefore the effect of start/end pitch should diminish when $n\textgreater1$. Third, even though feature weights are small for previous/next tones, in these results we didn't see an improvement when we exclude these features in the \texttt{no\_Ntone} baseline. Finally, we confirm the importance of the non-pitch linguistic features since the \texttt{no\_pitch} baseline significantly outperforms the \texttt{random} and the \texttt{MLE} baselines. 




\section{Discussion} \label{sec:discussion}

In this work, we first described a method to mine tone contour shape types from a large amount of tone data. These shape types, as well as those mined from tone $n$-grams with larger values of $n$ ($n\textgreater3$), can be further analyzed with linguistic knowledge to better understand the behaviors of tones in different tonal contexts in future works. We also showed that using linguistic and contextual factors, we can predict with reasonable accuracy the contour shape type that a tone $n$-gram will take in the MCPST data. We analyzed the feature importance to form a relative ranking of the linguistic factors. These results should be interpreted with caution because they are constrained by the particular representations of the linguistic features used in this study, as well as the accuracy of the NLP softwares used to extract them. Nonetheless, we envision that by mining correlates between speech prosody and automatic analysis of linguistic features extracted from the data, this line of work could have potential applications in improving the quality and naturalness of prosody in speech synthesis such as Text-To-Speech (TTS) technologies.

Previous works targeting information theory and information structure in prosody domain have largely looked at acoustic correlates directly, such as accent and duration, all of which may in turn have an impact on the shape of tone contours in speech production. Therefore, looking at tone contour shapes can be thought of as a different level of manifestation of such phenomena, an amalgamation of single dimension acoustic correlates (e.g., duration and intensity). It is also a level that is more difficult to quantify and measure in the traditional linguistic/phonetic investigations on a smaller scale. One possible extension of this work in the future is to look at specific ways tone contour shapes correlate with particular features when they are perturbed in a certain direction. Moreover, it is also of interest to demonstrate this change in a quantified manner using the information theory formulation.

In Section \ref{sec:related-work} we raised a fundamental theoretical question of whether there is a direct link between communicative functions and surface acoustic forms, a question that we found disagreement in literature (as summarized in \cite{Xu2005} and in Section \ref{sec:related-work} of the current paper). In this paper, we showed that by taking a data driven approach, we can predict the contour shape type of a prosodic category (such as a tone $n$-gram) using linguistic factors, even though we are not predicting its exact shape. In doing so, we give an approximate solution for a middle ground between the two theories.
\section*{Acknowledgements}
We thank Sankalp Gulati, Xavier Serra, Amir Zeldes, Elizabeth Zsiga, George Wilson, Richard Wright and Gina-Anne Levow for insights and discussions, as well as the anonymous reviewers for valuable comments on the previous versions of this paper.

\bibliography{naaclhlt2019}

\begin{thebibliography}{26}
\expandafter\ifx\csname natexlab\endcsname\relax\def\natexlab#1{#1}\fi

\bibitem[{Aylett and Turk(2004)}]{Aylett2004}
Matthew Aylett and Alice Turk. 2004.
\newblock \href {https://doi.org/10.1177/00238309040470010201} {{The smooth
  signal redundancy hypothesis: a functional explanation for relationships
  between redundancy, prosodic prominence, and duration in spontaneous
  speech.}}
\newblock \emph{Language and speech}, 47(Pt 1):31--56.

\bibitem[{Bard and Aylett(1999)}]{Bard99thedissociation}
E.~G. Bard and M.~P. Aylett. 1999.
\newblock The dissociation of deaccenting, givenness, and syntactic role in
  spontaneous speech.
\newblock In \emph{in ICPhs99}, pages 1753--1756.

\bibitem[{Blondel et~al.(2008)Blondel, Guillaume, Lambiotte, and
  Lefebvre}]{Blondel2008}
Vincent~D. Blondel, Jean-Loup Guillaume, Renaud Lambiotte, and Etienne
  Lefebvre. 2008.
\newblock \href {https://doi.org/10.1088/1742-5468/2008/10/P10008} {{Fast
  unfolding of communities in large networks}}.
\newblock \emph{Journal of Statistical Mechanics: Theory and Experiment},
  10008(10):6.

\bibitem[{Buring(2013)}]{buring2013}
Daniel Buring. 2013.
\newblock \href {https://doi.org/10.1017/CBO9780511804571.029} {\emph{Syntax,
  information structure, and prosody}}, Cambridge Handbooks in Language and
  Linguistics, pages 860--896. Cambridge University Press.

\bibitem[{Chen and Manning(2014)}]{DBLP:conf/emnlp/ChenM14}
Danqi Chen and Christopher~D. Manning. 2014.
\newblock \href {http://aclweb.org/anthology/D/D14/D14-1082.pdf} {A fast and
  accurate dependency parser using neural networks}.
\newblock In \emph{Proceedings of the 2014 Conference on Empirical Methods in
  Natural Language Processing, {EMNLP} 2014, October 25-29, 2014, Doha, Qatar,
  {A} meeting of SIGDAT, a Special Interest Group of the {ACL}}, pages
  740--750.

\bibitem[{Cooper et~al.(1985)Cooper, Eady, and Mueller}]{Cooper1985}
E~Cooper, J~Eady, and R~Mueller. 1985.
\newblock Acoustical aspects of contrastive stress in question-answer contexts.
\newblock \emph{The Journal of the Acoustical Society of America},
  77:2142--2156.

\bibitem[{Cooper and Sorenson(1981)}]{Cooper1981}
E~Cooper and M~Sorenson. 1981.
\newblock Fundamental frequency in sentence production.
\newblock \emph{New York: Spring-Verlag}.

\bibitem[{Gauthier et~al.(2007)Gauthier, Shi, and Xu}]{Gauthier2007}
Bruno Gauthier, Rushen Shi, and Yi~Xu. 2007.
\newblock \href {https://doi.org/10.1016/j.cognition.2006.03.002} {{Learning
  phonetic categories by tracking movements.}}
\newblock \emph{Cognition}, 103(1):80--106.

\bibitem[{Gulati et~al.(2016)Gulati, Serra, Ishwar, and Serra}]{Gulati16}
Sankalp Gulati, Joan Serra, Vignesh Ishwar, and Xavier Serra. 2016.
\newblock {Discovering raga motifs by characterizing communities in networks of
  melodic patterns}.
\newblock \emph{Proc. of IEEE International Conf. on Acoustics, Speech, and
  Signal Processing (ICASSP), pp. 286-290, Shanghai, China.}

\bibitem[{Guyon et~al.(2002)Guyon, Weston, Barnhill, and Vapnik}]{Guyon2002}
Isabelle Guyon, Jason Weston, Stephen Barnhill, and Vladimir Vapnik. 2002.
\newblock \href {https://doi.org/10.1023/A:1012487302797} {Gene selection for
  cancer classification using support vector machines}.
\newblock \emph{Machine Learning}, 46(1):389--422.

\bibitem[{Ladd(1996)}]{Ladd1996}
R~Ladd. 1996.
\newblock Intonational phonology.
\newblock \emph{Cambridge University Press, Cambridge}.

\bibitem[{Levow(2005)}]{Levow2005}
Gina-Anne Levow. 2005.
\newblock {Context in multi-lingual tone and pitch accent recognition.}
\newblock \emph{Interspeech}, pages 1809--1812.

\bibitem[{Li(2009)}]{KLi09}
Kening Li. 2009.
\newblock The information structure of mandarin chinese: Syntax and prosody.
\newblock \emph{PhD Dissertation, Department of Linguistics, University of
  Washington}.

\bibitem[{Liberman and Pierrehumbert(1984)}]{Liberman1984}
M~Liberman and J~Pierrehumbert. 1984.
\newblock Intonational invariance under changes in pitch range and length.
\newblock \emph{M. Aronoff and R. Oehrle (Eds.), Language Sound Structure.
  M.I.T. Press, Cambridge, Massachusetts}, pages 157--233.

\bibitem[{Liu et~al.(2006)Liu, Surendran, and Xu}]{Liu2006}
Fang Liu, Dinoj Surendran, and Yi~Xu. 2006.
\newblock Classification of statement and question intonation in mandarin.
\newblock In \emph{Proceedings of Speech Prosody}.

\bibitem[{Manning et~al.(2014)Manning, Surdeanu, Bauer, Finkel, Bethard, and
  McClosky}]{manning-EtAl:2014:P14-5}
Christopher~D. Manning, Mihai Surdeanu, John Bauer, Jenny Finkel, Steven~J.
  Bethard, and David McClosky. 2014.
\newblock \href {http://www.aclweb.org/anthology/P/P14/P14-5010} {The
  {Stanford} {CoreNLP} natural language processing toolkit}.
\newblock In \emph{Association for Computational Linguistics (ACL) System
  Demonstrations}, pages 55--60.

\bibitem[{Prom-on et~al.(2009)Prom-on, Xu, and Thipakorn}]{Prom-on2009}
Santitham Prom-on, Yi~Xu, and Bundit Thipakorn. 2009.
\newblock \href {https://doi.org/10.1121/1.3037222} {{Modeling tone and
  intonation in Mandarin and English as a process of target approximation.}}
\newblock \emph{The Journal of the Acoustical Society of America},
  125(1):405--24.

\bibitem[{Recasens et~al.(2013)Recasens, de~Marneffe, and
  Potts}]{recasens-etal-2013-life}
Marta Recasens, Marie-Catherine de~Marneffe, and Christopher Potts. 2013.
\newblock \href {https://www.aclweb.org/anthology/N13-1071} {The life and death
  of discourse entities: Identifying singleton mentions}.
\newblock In \emph{Proceedings of the 2013 Conference of the North American
  Chapter of the Association for Computational Linguistics: Human Language
  Technologies}, pages 627--633, Atlanta, Georgia. Association for
  Computational Linguistics.

\bibitem[{R{\"{o}}esiger and Riester(2015)}]{Roesiger2015}
Ina R{\"{o}}esiger and Arndt Riester. 2015.
\newblock {Using prosodic annotations to improve coreference resolution of
  spoken text}.
\newblock In \emph{Proceedings of the 53rd Annual Meeting of the Association
  for Computational Linguistics and the 7th International Joint Conference on
  Natural Language Processing}, volume~2.

\bibitem[{Surendran(2007)}]{Surendran2007}
Dinoj Surendran. 2007.
\newblock {Analysis and Automatic Recognition of Tones in Mandarin Chinese}.
\newblock \emph{PhD Thesis, Department of Computer Science, University of
  Chicago}.

\bibitem[{Wang and Xu(2011)}]{Wang2011}
Bei Wang and Yi~Xu. 2011.
\newblock \href {https://doi.org/10.1016/j.wocn.2011.03.006} {{Differential
  prosodic encoding of topic and focus in sentence-initial position in Mandarin
  Chinese}}.
\newblock \emph{Journal of Phonetics}, 39(4):595--611.

\bibitem[{Xu(1997)}]{Xu1997}
Yi~Xu. 1997.
\newblock \href {https://doi.org/10.1006/jpho.1996.0034} {{Contextual tonal
  variations in Mandarin}}.
\newblock \emph{Journal of Phonetics}, 25(1):61--83.

\bibitem[{Xu(2005)}]{Xu2005}
Yi~Xu. 2005.
\newblock \href {https://doi.org/10.1016/j.specom.2005.02.014} {{Speech melody
  as articulatorily implemented communicative functions}}.
\newblock \emph{Speech Communication}, 46(3-4):220--251.

\bibitem[{Xu et~al.(2004)Xu, Xu, Sun, Laboratories, and Haven}]{Xu2004}
Yi~Xu, Ching~X Xu, Xuejing Sun, Haskins Laboratories, and New Haven. 2004.
\newblock {On the Temporal Domain of Focus}.
\newblock \emph{Speech Prosody}, pages 81--84.

\bibitem[{Yu(2011)}]{KYu11}
Kristine Yu. 2011.
\newblock The learnability of tones from the speech signal.
\newblock \emph{PhD Dissertation, Department of Linguistics, UCLA}.

\bibitem[{Zhang(2016)}]{W16-2001}
Shuo Zhang. 2016.
\newblock \href {https://doi.org/10.18653/v1/W16-2001} {Mining linguistic tone
  patterns with symbolic representation}.
\newblock In \emph{Proceedings of the 14th SIGMORPHON Workshop on Computational
  Research in Phonetics, Phonology, and Morphology}, pages 1--9. Association
  for Computational Linguistics.

\end{thebibliography}
\bibliographystyle{acl_natbib}



\end{document}